\documentclass[11pt,a4paper]{article}
\usepackage[hyperref]{acl2019}
\usepackage{times}
\usepackage{latexsym}
\usepackage{multirow}
\usepackage{enumitem}
\usepackage{graphicx}
\usepackage[english]{babel} 
\usepackage{url}
\usepackage{latexsym}
\usepackage{helvet}
\usepackage{courier}
\usepackage{subfigure}
\usepackage{multirow}
\usepackage{indentfirst}
\usepackage{bm}
\usepackage{amsfonts}
\usepackage{amsmath}
\usepackage{float}
\usepackage{amsthm}
\usepackage{amsmath}
\usepackage{booktabs}
\usepackage{tikz}

\aclfinalcopy 

\newtheorem{myDef}{Definition} 

\title{ELG: An Event Logic Graph}

\author{Xiao Ding, Zhongyang Li, Ting Liu\thanks{~~Corresponding Author}, Kuo Liao \\
	Research Center for Social Computing and Information Retrieval \\
	Harbin Institute of Technology, Harbin, 150001, China \\
	{\tt \{xding, zyli, tliu, kliao\}@ir.hit.edu.cn}\\
}

\date{}

\begin{document}
\maketitle
\begin{abstract}
  The evolution and development of events have their own basic principles, which make events happen sequentially. Therefore, the discovery of such evolutionary patterns among events are of great value for event prediction, decision-making and scenario design of dialog systems. However, conventional knowledge graph mainly focuses on the entities and their relations, which neglects the real world events. In this paper, we present a novel type of knowledge base --- \textbf{Event Logic Graph} (ELG), which can reveal evolutionary patterns and development logics of real world events. Specifically, ELG is a directed cyclic graph, whose nodes are events, and edges stand for the sequential, causal, conditional or hypernym\text{-}hyponym (``is-a'') relations between events. We constructed two domain ELG: financial domain ELG, which consists of more than 1.5 million of event nodes and more than 1.8 million of directed edges, and travel domain ELG, which consists of about 30 thousand of event nodes and more than 234 thousand of directed edges. Experimental results show that ELG is effective for the task of script event prediction.
\end{abstract}

\section{Introduction}

\noindent The evolution and development of events have their own underlying principles, leading to events happen sequentially. For example, the sentence ``\emph{After having lunch, Tom paid the bill and left the restaurant}'' shows a sequence of event evolutions: ``\emph{have lunch}''$\rightarrow$``\emph{pay the bill}''$\rightarrow$``\emph{leave the restaurant}''. This event sequence is a common pattern for the scenario of having lunch in a restaurant. Such patterns can reveal the basic rules of event evolutions and human behaviors. Hence, it is of great value for many Artificial Intelligence (AI) applications, such as discourse understanding, intention identification and dialog generation.

However, traditional knowledge graph takes the entity as the research focus, and investigate the properties of entities and the relationships between entities, which lacks of event-related knowledge. In order to discover the evolutionary patterns and logics of events, we propose an event-centric knowledge graph --- \textbf{Event Logic Graph} (ELG) and the framework to construct ELG. ELG is a directed cyclic graph, whose nodes are events, and edges stand for the sequential (the same meaning with ``temporal''), causal, conditional or hypernym\text{-}hyponym (``is-a'') relations between events. Essentially, ELG is an event logic knowledge base, which can reveal evolutionary patterns and development logics of real world events. 

To construct ELG, the first step is to extract events from raw texts, and then recognize the sequential, causal, conditional or hypernym\text{-}hyponym relations between two events and distinguish the direction of each sequential or causal relation. In the end, we need to merge event pairs to obtain the final event logic graph by connecting semantically similar events and generalizing each specific event.

\begin{figure*}[!tb]
	\centering
	\includegraphics[width=0.8\textwidth]{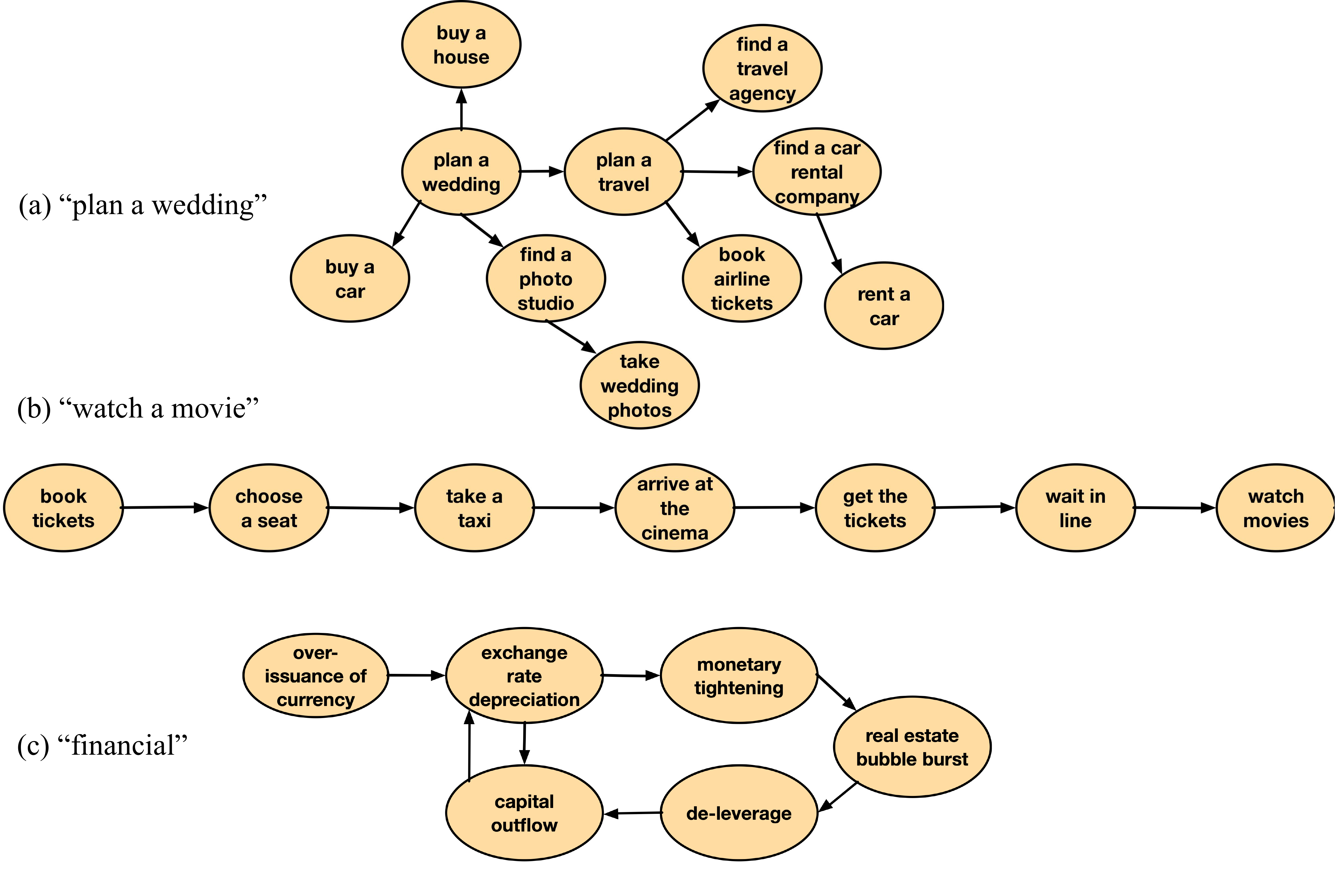}
	\caption{Tree structured event logic graph under the scenario of ``plan a wedding'', chain structured event logic graph under the scenario of ``watch a movie'' and cyclic structured event logic graph from the ``financial'' domain.}
	\label{fig:eegstructure}
\end{figure*}

Numerous efforts have been made to extract temporal and causal relations from texts. As the most commonly used corpus, TimeBank~\cite{pustejovsky2003timebank} has been adopted in many temporal relation extraction studies. Mani et al.~\shortcite{mani2006machine} applied the temporal transitivity rule to greatly expand the corpus. Chambers et al.,~\shortcite{chambers2007classifying} used previously learned event attributes to classify the temporal relationship. For causality relation extraction, Zhao et al.~\shortcite{zhao2017constructing} extracted multiple kinds of features to recognize causal relations between two events in the same sentence. Radinsky et al.~\shortcite{radinsky2012learning} automatically extracted cause-effect pairs from large quantities of news headlines by designed causal templates for predicting news events. However, most of these studies only extract temporal or causal event pairs from single sentences, which are discrete knowledge pieces, and fail to organize them into a knowledge graph. In this paper, we further propose to organize the universal event evolutionary principles and patterns into a knowledge base based on the extracted temporal and causal event pairs.


The main contributions of this paper are threefold. \textbf{First}, we are among the first to propose the definition of ELG. \textbf{Second}, we propose a promising framework to construct ELG from a large-scale unstructured text corpus. \textbf{Third}, experimental results show that ELG is capable of improving the performances of downstream applications, such as script event prediction.

\section{Event Logic Graph}


\begin{myDef}
	ELG is a directed cyclic graph, whose nodes are events, and edges stand for the sequential, causal, conditional or hypernym\text{-}hyponym relations between events. Essentially, ELG is an event logic knowledge base, which reveals evolutionary patterns and development logics of real world events.
\end{myDef}

Figure~\ref{fig:eegstructure} demonstrates three different event logic subgraphs of three different scenarios. Concretely, Figure~\ref{fig:eegstructure}~(a) describes a sequence of events under the scenario of ``plan a wedding'', which usually happen in the real world, and have evolved into the fixed human behavior patterns. For example, ``plan a wedding'' usually follows by ``buy a house'', ``buy a car'' and ``plan a travel''. This kind of commonsense event evolutionary patterns are usually hidden behind human beings' daily activities, or online user generated contents. To the best of our knowledge, this kind of commonsense knowledge is not explicitly stored in any existing knowledge bases, so that, we propose constructing an ELG to store it. 


\begin{figure*}[!tb]
	\centering	
	\includegraphics[width=0.9\textwidth]{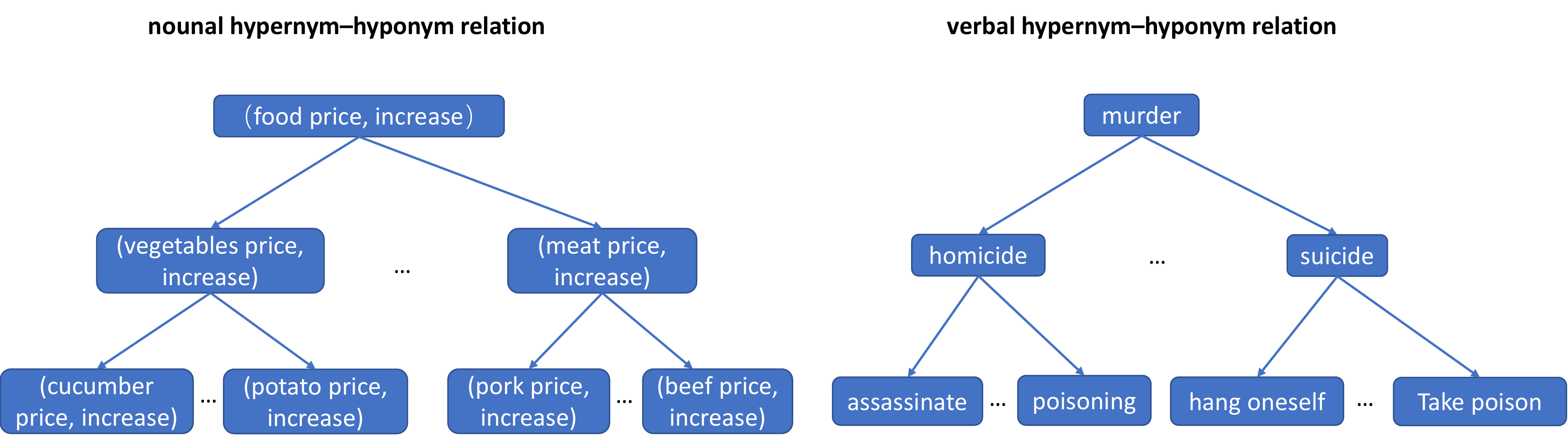}	
	\caption{The examples of two kinds of hypernym\text{-}hyponym relations in ELG.}
	\label{fig:shangxia}
\end{figure*}

\begin{table*} [!tb]\small
	\centering
	\begin{tabular}{l p{5cm} p{3.5cm}}
		\toprule
		\textbf{Aspects}&\textbf{Event Logic Graph} & \textbf{Knowledge Graph} \\
		\hline
		Research Subject & events and their relations & entities and their relations\\
		Structure & directed and cycled graph & directed and cycled graph \\
		Forms of Knowledge  & sequential, causal, conditional and hypernym\text{-}hyponym relations between events & entities' attributes and their relations \\
		Certainty of Knowledge & mostly probabilistic & mostly deterministic \\
		\bottomrule
	\end{tabular}
	\caption{Comparison between Event Logic Graph and Knowledge Graph.}
	\label{tab:compare}
\end{table*}

In ELG, events are represented as \textbf{abstract, generalized and semantic complete} event tuples $E$ = ($S, P, O$), where $P$ is the action, $S$ is the actor and $O$ is the object on which the action is performed. In our definition, each event must contain a trigger word (i.e., $P$), such as ``run'', which mainly indicates the type of the event. In different application scenarios, $S$ and $O$ can be omitted, respectively, such as (\emph{watch, a movie}) and (\emph{climate, warming}). In general, events and the degree of abstraction of an event are closely related to the scene in which the event occurred, and a single event may become too abstract to understand without the context scenario.

\textbf{Abstract and generalized} means that we do not concern about the exact participants, location and time of an event. For example, ``who watch movies'' and ``watch which movie'' are not important in ELG. \textbf{Semantic complete} means that human beings can understand the meaning of the event without ambiguity. For example, (\emph{have, a hot pot}) and (\emph{go to, the airport}) are reasonable event tuples to represent events. While (\emph{go, somewhere}), (\emph{do, the things}) and (\emph{eat}) are unreasonable or incomplete event representations, as their semantics are too vague to be understood.

We have four categories of directed edges: sequential directed edges, causal directed edges, conditional directed edges and hypernym\text{-}hyponym directed edges, which indicate different relationships between events. The sequential relation between two events refers to their partial temporal orderings. For example, given the sentence ``After having lunch, Tom paid the bill and left the restaurant.'' (\emph{have, lunch}), (\emph{pay, the bill}) and (\emph{leave the restaurant}) compose a sequential event chain.

The causal relation is the relation between the cause event and the effect event. For example, given the sentence ``The nuclear leak in Japan leads to serious ocean pollution.'' (\emph{nuclear, leak}) is the cause event, and (\emph{ocean, pollution}) is the effect event. It is obvious that the causal relation must be sequential.

The conditional relation is a logical relation in which the illocutionary act employing one of a pair of propositions is expressed or implied to be true or in force if the other proposition is true. For example, ``\emph{study hard}'' is the condition of ``\emph{achieve good performances}''.

The hypernym\text{-}hyponym relation is a kind of ``is-a'' relationship between events, which includes two different types of hypernym\text{-}hyponym relations: nounal hypernym\text{-}hyponym relation and verbal hypernym\text{-}hyponym relation, as shown in Figure~\ref{fig:shangxia}.

ELG is different from traditional Knowledge Graph in many aspects. As shown in Table~\ref{tab:compare}, ELG focuses on events, and the directed edges between event nodes stand for the sequential, causal or hypernym\text{-}hyponym relations between them. The sequential and causal relations in ELG are probabilistic. In contrast, traditional Knowledge Graph focuses on entities, and its edges stand for the attributes of entities or relations between them. There are usually thousands of types of relations between entities. Moreover, the attributes and relations in Knowledge Graph are mostly deterministic.

\section{Architecture}
\begin{figure}[!tb]
	\centering
	
	\includegraphics[width=0.5\textwidth]{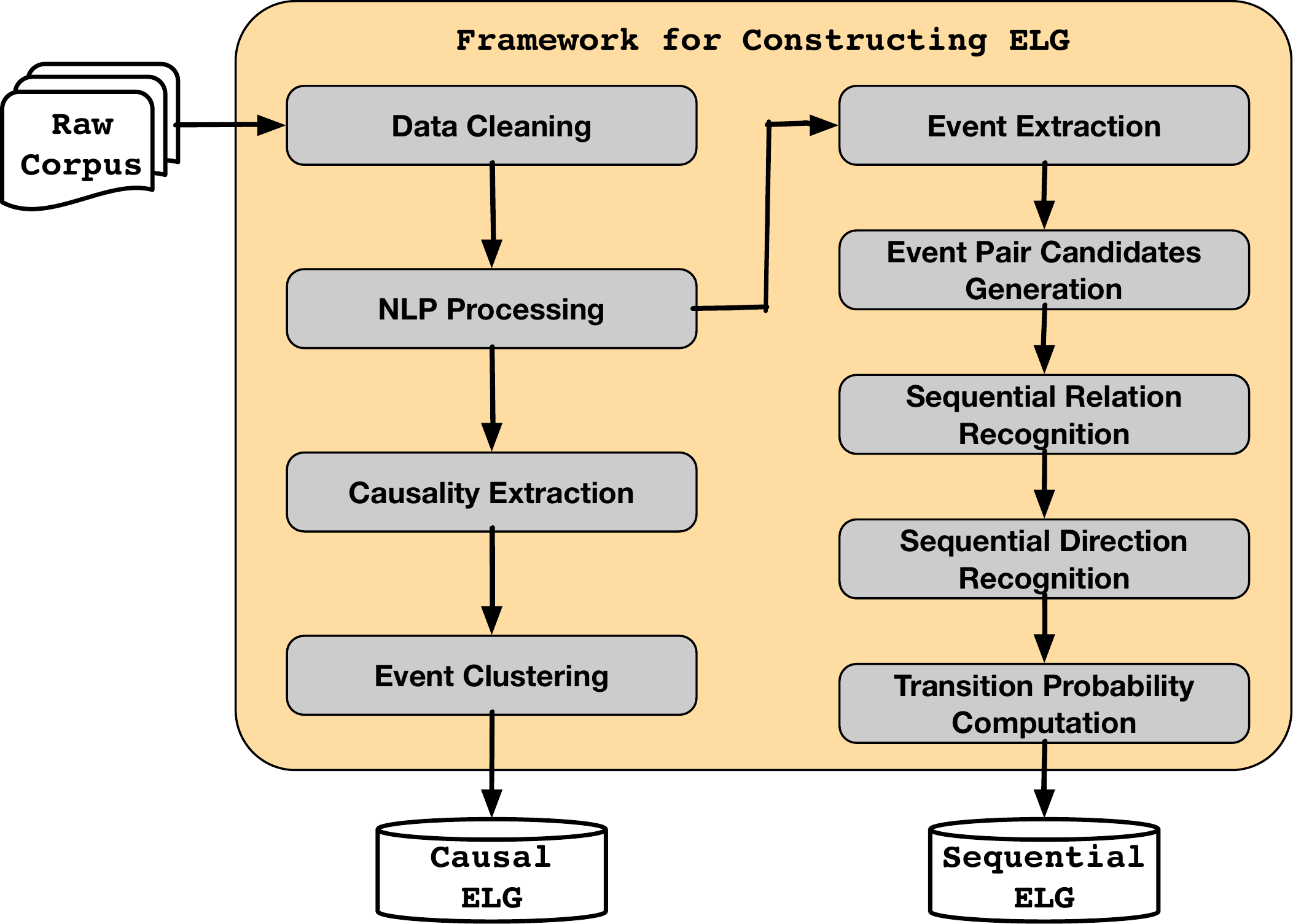}
	
	\caption{Out proposed framework for constructing ELG from large-scale unstructured texts.}
	\label{fig:pipeline}
\end{figure}

\noindent As illustrated in Figure~\ref{fig:pipeline}, we propose a framework to construct an ELG from large-scale unstructured texts, including data cleaning, natural language processing, event extraction, sequential relation and direction classification, causality extraction and transition probability computation. 


After cleaning the data, a series of natural language processing steps including segmentation, part-of-speech tagging, and dependency parsing are conducted for event extraction. Tools provided by Language Technology Platform (LTP)~\cite{che2010ltp} are used for this preprocessing.


\subsection{Open Event Extraction}

\noindent We extract structured events from free text using Open IE technology and dependency parsing. Given a sentence obtained from texts, we first adopt the event extraction methods described in~\cite{ding2013building} to extract the candidate tuples of the event $(S, P, O, X)$, and then parse the sentence with LTP~\cite{che2010ltp} to extract the subject, object and predicate. 

We filter out the low-frequency event tuples by a proper threshold, to exclude the event tuples extracted due to segmentation and dependency parsing errors. Some too general events such as ``go somewhere'' and ``do something'' are removed by regular expressions with a dictionary.



\begin{table} [!tb]\small
	\centering
	\begin{tabular}{p{4cm}|p{2.9cm}}
		\hline
		\textbf{Frequency-based Features} & \textbf{Ratio-based Features} \\
		\hline
		T1: count of (A, B) &R1: T2/T1 \\
		T2: count of (A, B) where A occurs before B & R2: T1/T4 \newline  R3: T1/T5  \\
		T3: count of (A, B) where B occurs before A &  R4: T1/T6 \newline R5: T1/T7 \\
		T4: count of A &  R6: T1/T8\\
		T5: count of B &  R7: T1/T9\\
		T6: count of verb-A & R8: T6/T4\\
		T7: count of object-A & R9: T7/T4\\
		T8: count of verb-B & R10: T8/T5\\
		T9: count of object-B & R11: T9/T5\\
		
		\hline
		
		\textbf{Context-based Features} & \textbf{PMI-based Features} \\
		\hline
		C1: the number of contexts in which A and B co-occur & A1: PMI of verb-A and verb-B\\
		C2: average length of C1 & A2: PMI of A and B\\
		C3: word embeddings of contexts in which A and B co-occur & A3: PMI of verb-A and object-B\\
		C4: the postag of contexts in which A and B co-occur & A4: PMI of object-A and verb-B \\
		C5: phrase embeddings of A and B & A5: PMI of object-A and object-B\\
		\hline
	\end{tabular}
\caption{The features used for sequential relation and direction classification.}
	\label{tab:features}
\end{table}

\subsection{Sequential Relation and Direction Recognition}

\noindent Given an event pair candidate (A, B), sequential relation recognition is to judge whether it has a sequential relation or not. For the ones having sequential relations, direction recognition should be conducted to distinguish the direction. For example, we need to recognize that there is a directed edge from (\emph{buy, tickets}) to (\emph{watch, a movie}). We regard the sequential relation and direction recognition as two separate binary classification tasks. 

As shown in Table~\ref{tab:features}, multiple kinds of features are extracted for these two supervised classification tasks. Details about the intuition why we choose these features are described below:

\begin{itemize}[leftmargin=*]
	\item \textbf{Frequency-based Features:} For a candidate event pair (A, B), the frequency-based features include their co-occur frequency (T1 to T3), respective frequency of A and B in the whole corpus (T4 and T5), and respective frequency of each event argument (T6 to T9).
	
	\item \textbf{Ratio-based Features:} Some meaningful combinations between frequency-based features may provide extra information that is useful for sequential relation and direction classification, shown in Table~\ref{tab:features} (R1 to R11).
	
	\item \textbf{Context-based Features:} We believe that the contexts of event A and B are important features for identifying their sequential relation. We devise context-based features that capture the contextual semantic information of A and B, shown in Table~\ref{tab:features} (C1 to C5). 
	
	\item \textbf{PMI-based Features:} We also investigate the pointwise mutual information (PMI) between event A and B, shown in Table~\ref{tab:features} (A1 to A5). 
\end{itemize}

\noindent \textbf{Transition Probability Computation} Given an event pair (A, B), we use the following equation to approximate the transition probability from the event A to the event B:

\begin{equation} 
P(\text{B}|\text{A})= \frac {f(\text{A},\text{B})}{f(\text{A})},
\end{equation}
where $f(\text{A},\text{B})$ is the co-occurrence frequency of event pair (A, B), and $f(\text{A})$ is the frequency of event A in the whole corpus.


\subsection{Causality Extraction}
\subsubsection{Unsupervised Causality Extraction}
\noindent The first step to construct causal relation ELG is to identify cause-effect pairs from unstructured natural language texts. As the amount of data is extremely large (millions of documents), obtaining human-annotated pairs is impossible. We find that causal relations expressed in text have various forms. We therefore provide a procedure similar to our previous work \cite{zhao2017constructing}, which can automatically identify mentions of causal events from natural language texts.

We construct a set of rules to extract mentions of causal events. Each rule follows the template of $<$\textbf{Pattern, Constraint, Priority}$>$, where \textbf{Pattern} is a regular expression containing a selected connector, \textbf{Constraint} is a syntactic constraint on sentences to which the pattern can be applied, and \textbf{Priority} is the priority of the rule if several rules are matched. For example, we use the pattern ``[\emph{cause}] leads to [\emph{effect}]'' to extract the causal relation between two events.



	
	
	
	
	
	

\subsubsection{Supervised Causality Extraction}
\begin{figure} [!tb]
	\centering
	\includegraphics[width=0.55\textwidth]{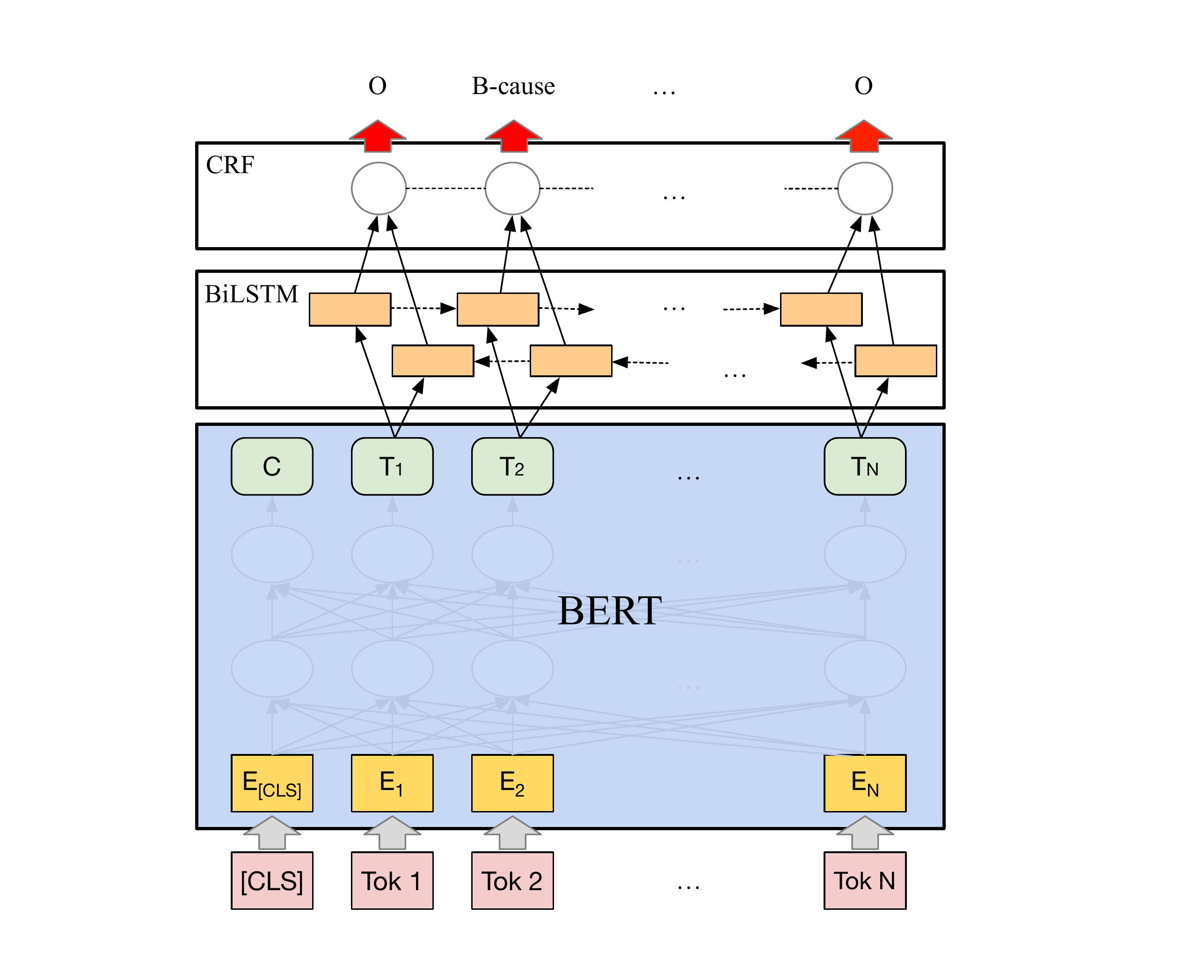}
	\caption{Architecture of the supervised causality extraction model.}
	\label{fig:bert}
\end{figure}

\noindent As illustrated in Figure~\ref{fig:bert}, we also use Bert and BiLSTM+CRF model to extract causal relations. Language model pre-training has shown to be very effective for learning universal language representations by leveraging large amounts of unlabeled data. Some of the most prominent models are ELMo \cite{peters2018deep}, GPT \cite{Radford2018GPT}, and BERT \cite{devlin2019bert:}. BERT uses the bidirectional transformer architecture. There are two existing strategies for applying pre-trained language models to downstream tasks: feature-based and fine-tuning.

In this paper, we annotate each token in a sentence with following tags: B-cause, I-cause, B-effect, I-effect and O.
The tag ``B-cause'' refers to the beginning token of the cause event and
each rest token in the cause event is represented by ``I-cause".
The tag ``O'' refers to the normal token which is irrelevant with causality.
We feed the hidden representation $T_i$ for each token $i$ as the input layer of BiLSTM.
These hidden representations $T_i$ can be viewed as semantic features learnt from Bert model.
The output representation layer of BiLSTM is then fed into the classification layer to predict the causal tags.
The predictions in the classification layer are conditioned on the surrounding predictions by using the CRF method.

\begin{figure}[!tb]
	\centering
	
	\includegraphics[width=0.3\textwidth]{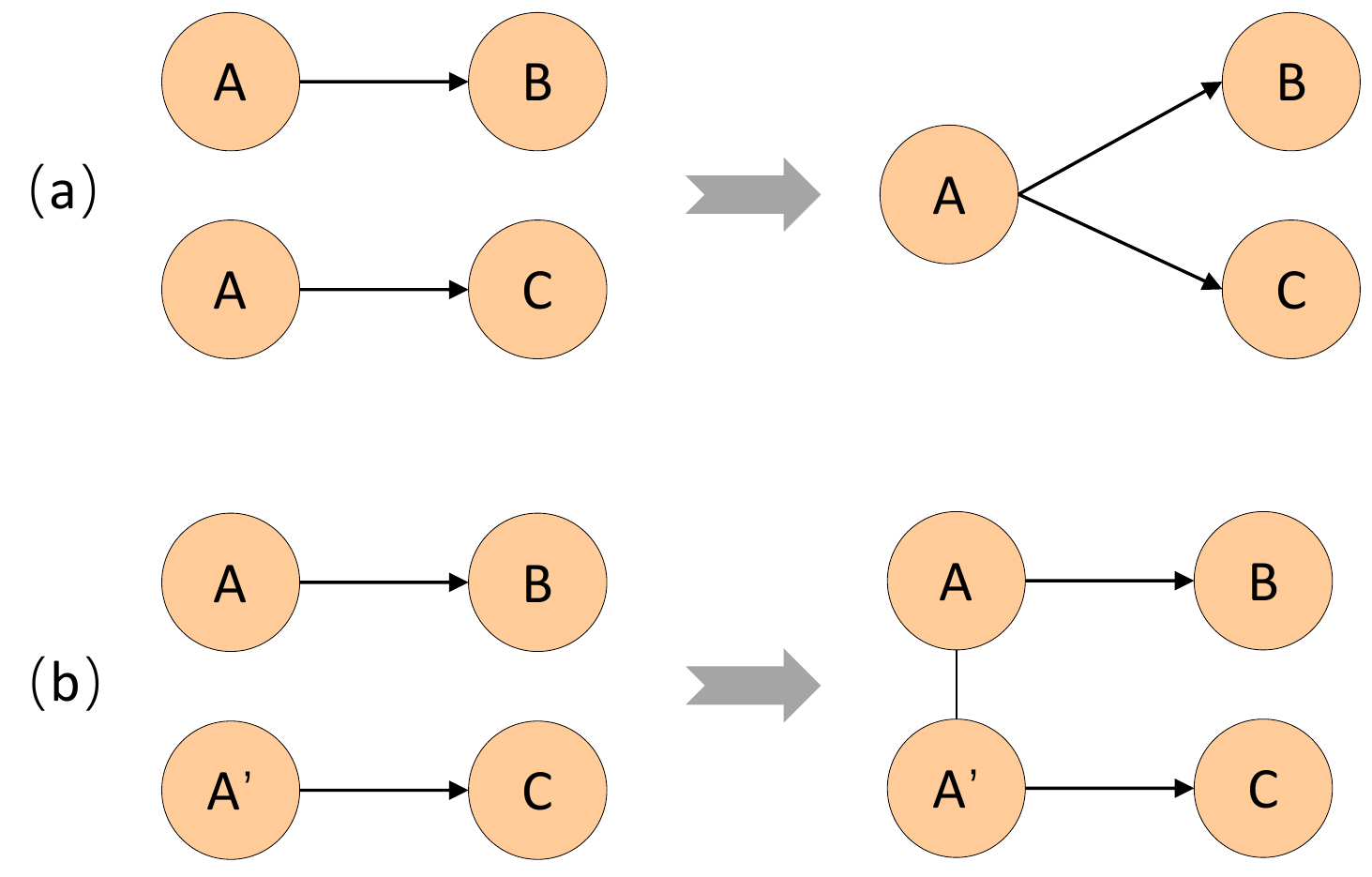}
	
	\caption{Connecting event pairs to form the graph structure.}
	\label{fig:graph}
\end{figure}

\subsection{Event Generalization}

\noindent Given large amount of event pairs extracted in previous steps, we need to connect event pairs to form a graph structure. Intuitively, as shown in Figure~\ref{fig:graph}~(a), if we can find the same event in two event pairs, it is easy to form the graph structure. However, as the extracted events are discretely represented by bag-of-words, we can hardly find two identical events. Hence, as shown in Figure~\ref{fig:graph}~(b), we propose to find the semantically similar events (A and A') and connect them. To this end, we propose learning distributed representations for each event, and utilize the cosine similarity to measure the semantic similarity between two event vectors. We use the framework of neural tensor networks to learn event embeddings, as described in our previous work \cite{ding2015deep}.

\begin{figure} [!tb]
	\centering
	\includegraphics[width=1\columnwidth]{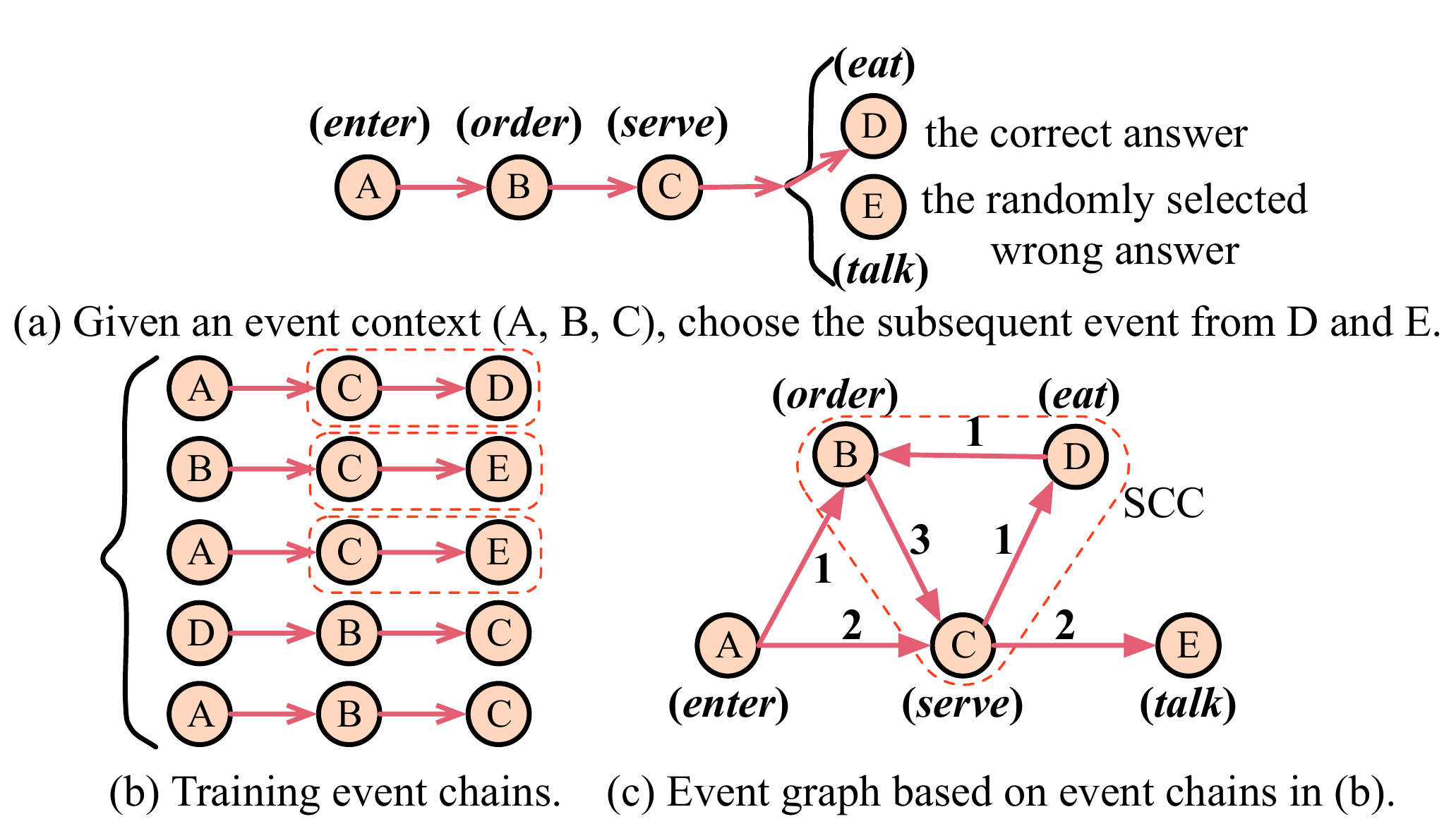}
	\caption{In (b), the training event chains show that C and E have a stronger relation than C and D, therefore event pair-based and chain-based models are very likely to choose the wrong, random candidate E. In (c), events B, C, and D compose a strongly connected component. This special graph structure contains dense connections information, which can help learn better event representations for choosing the correct subsequent event D.}
	\label{fig:example}
\end{figure}

\subsection{Application of the ELG}
\noindent We investigate the application of the ELG on the task of script event prediction~\cite{li2018constructing}. Figure~\ref{fig:example}~(a) gives an example to motive our idea of using ELG. Given an event context A(\textit{enter}), B(\textit{order}), C(\textit{serve}), we need to choose the most reasonable subsequent event from the candidate list D(\textit{eat}) and E(\textit{talk}), where D(\textit{eat}) is the correct answer and E(\textit{talk}) is a randomly selected candidate event that occurs frequently in various scenarios. Pair-based and chain-based models trained on event chains datasets (as shown in Figure~\ref{fig:example}~(b)) are very likely to choose the wrong answer E, as the training data shows that C and E have a stronger relation than C and D. As shown in Figure~\ref{fig:example}~(c), by constructing an ELG based on training events, context events B, C and the candidate event D compose a strongly connected component, which indicates that D is a more reasonable subsequent event, given context events A, B, C.

Based on the ELG and our proposed scaled graph neural network (SGNN), we achieved the state-of-the-art performance of script event prediction in the work of~\cite{li2018constructing}. We can also incorporate ELG into dialog systems to ensure that the auto-reply answers are more logical. 

\begin{figure}[!tb]
	\centering
	
	\includegraphics[width=0.5\textwidth]{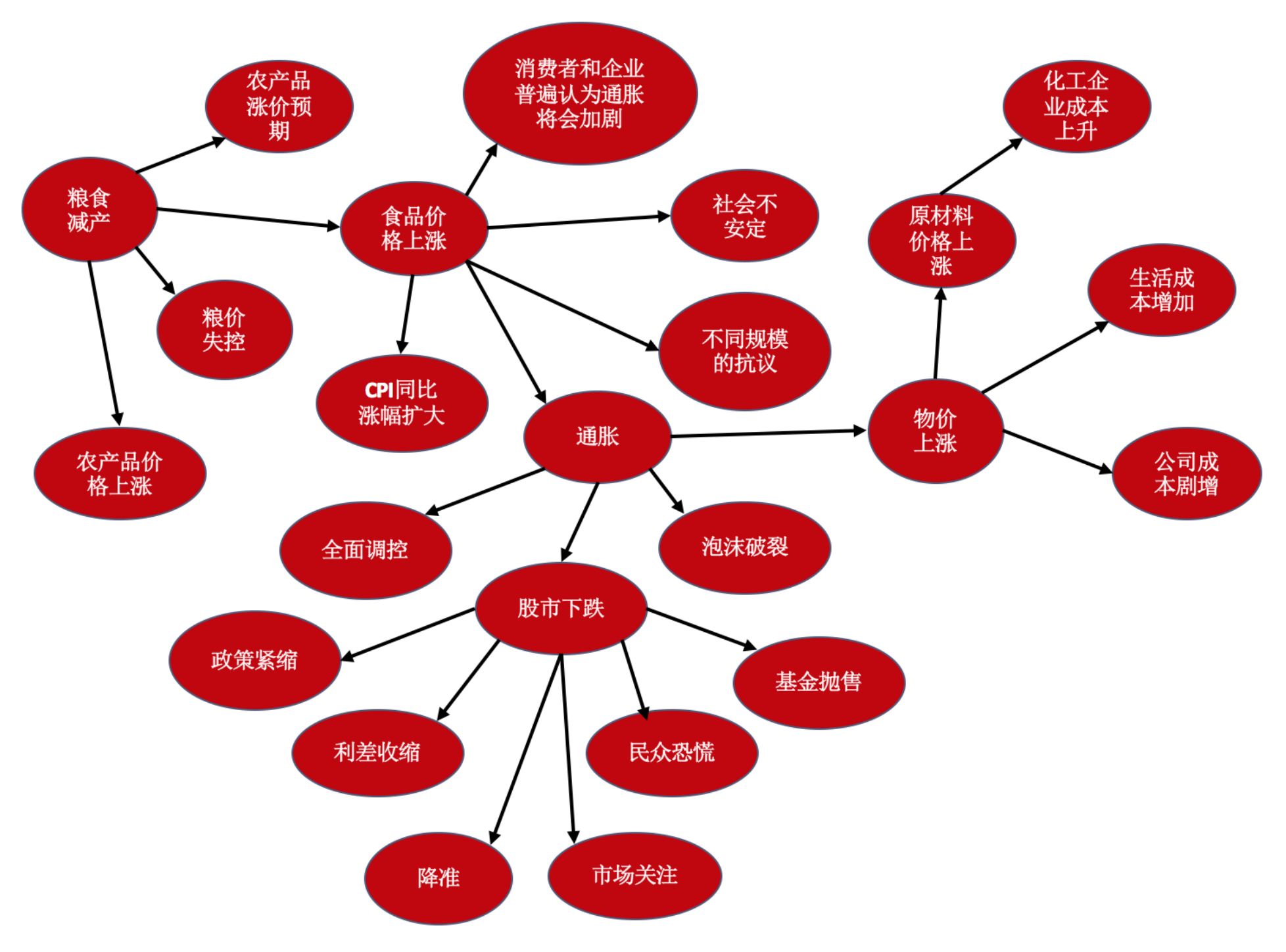}
	
	\caption{A screenshot of the online ELG demo.}
	\label{fig:eeg}
\end{figure}

Moreover, we constructed a financial ELG, which consists of more than 1.5 million of event nodes and 1.8 million of directed edges. Figure~\ref{fig:eeg} shows a screenshot of the online financial ELG (http://elg.8wss.com/). The user can enter any financial event in the search box, such as ``inflation''. Our demo can generate an event logic graph with ``inflation'' as the initial node. The red node in Figure~\ref{fig:eeg} is the input event; the yellow nodes are evolutionary nodes according to the input event, and the green nodes are semantically similar events with yellow nodes. We also give the extracted cause event, effect event and contexts in the right column to help users better understand the graph. Based on this financial ELG and the current events, we can infer what events will happen in the future.

\section{Experiments}

\noindent In this section, we conduct three kinds of experiments. First, we recognize whether two events has a sequential relation and its direction. Second, we extract casual relations between events based on our proposed unsupervised and supervised approaches. Finally, we use the downstream task: script event prediction to show the effectiveness of ELG. 

\begin{table} [tbp]\small
    \centering
	\begin{tabular*}{0.47\textwidth}{l c c c}
	\toprule
        &\textbf{Total} & \textbf{Positive} & \textbf{Negative} \\\hline
        Sequential Relation  & 2,173 & 1,563 & 610 \\
        Sequential Direction & 1,563 & 1,349 & 214 \\
    \bottomrule
    \end{tabular*}
    \caption{The detailed data statistics.}
    \label{tab:data_set}
\end{table}

\subsection{Dataset Description}

\noindent We annotated 2,173 event pairs with high co-occurrence frequency ($\geq 5$) from the dataset. Each event pair (A, B) is ordered that A occurs before B with a higher frequency than B occurs before A. In the annotation process, the annotators are provided with the event pairs and their corresponding contexts. They need to judge whether there is a sequential relation between two events from a commonsense perspective. If true, they also need to give the sequential direction. For example, ``watch movies'' and ``listen to music'' are tagged as no sequential relation (negative), while ``go to the railway station'' and ``by tickets'' are tagged as having a  sequential relation (positive), and the sequential direction is from the former to the latter (positive). The detailed statistics of our dataset are listed in Table~\ref{tab:data_set}. The positive and negative examples are very imbalanced. So we over sample the negative examples in training set to ensure the number of positive and negative training examples are equal. 

For causal relation experiment, we crawled 1,362,345 Chinese financial news documents from online websites, such as Tencent\footnote{http://finance.qq.com/} and Netease\footnote{http://money.163.com/}. All the headlines and main contents are exploited as the experiment corpus. We manually annotated 1,000 sentences to evaluate the causality extraction performance.

Script event prediction \cite{chambers2008narrative} is a challenging event-based commonsense reasoning task, which is defined as giving an existing event context, one needs to choose the most reasonable subsequent event from a candidate list. Following \citeauthor{wang2017integrating} (\citeyear{wang2017integrating}), we evaluate on the standard multiple choice narrative cloze (MCNC) dataset \cite{MarkGW-AAAI16}.

\subsection{Baselines and Evaluation Metrics}

\subsubsection{Sequential Relation}
\noindent For sequential relation recognition, PMI score of an event pair is used as the baseline method. For sequential direction recognition, if event A occurs before B with a higher frequency than B occurs before A, we regard the sequential direction as from event A to event B. This is called the \textbf{Preceding Assumption}, which is used as the baseline method for sequential direction recognition.

For sequential relation and direction recognition, four classifiers are used for these classification tasks, which are naive Bayes classifier (NB), logistic regression (LR), multiple layer perceptron (MLP) and support vector machines (SVM). We explored different feature combinations to find the best feature set for both classification tasks. All experiments are conducted using five-fold cross validation. The final experiment result is the average performance of ten times of implementations.

Two kinds of evaluation metrics are used to evaluate the performance of our proposed methods: accuracy and F1 metric.

\subsubsection{Causal Relation}
\noindent For causal relation mining, we mainly conduct experiments to evaluate the causality extraction system. The same evaluation metrics as in sequential relation experiment are used to evaluate the performance of causality extraction. We mainly compare unsupervised rule-based causality extraction approach with supervised bert-based causality extraction approach.

\subsubsection{Script Event Prediction}
\noindent We compare our model with the following baseline methods, and follow previous work \cite{wang2017integrating} using the accuracy as the evaluation metric.
\begin{itemize}[leftmargin=*]
	\item \textbf{PMI} \cite{chambers2008narrative} is the co-occurrence-based model that calculates predicate-GR event pairs relations based on Pairwise Mutual Information.
	\item \textbf{Bigram} \cite{jans2012skip} is the counting-based skip-grams model that calculates event pair relations based on bigram probabilities.
	\item \textbf{Word2vec} \cite{mikolov2013distributed} is the widely used model that learns word embeddings from large-scale text corpora. The learned embeddings for verbs and arguments are used to compute pairwise event relatedness scores.
	\item \textbf{DeepWalk} \cite{perozzi2014deepwalk} is the unsupervised model that extends the word2vec algorithm to learn embeddings for networks.
	\item \textbf{EventComp} \cite{MarkGW-AAAI16} is the neural network model that simultaneously learns embeddings for the event verb and arguments, a function to compose the embeddings into a representation of the event, and a coherence function to predict the strength of association between two events.
	\item \textbf{PairLSTM} \cite{wang2017integrating} is the model that integrates event order information and pairwise event relations together by calculating pairwise event relatedness scores using the LSTM hidden states as event representations.
\end{itemize}

\subsection{Results and Analysis}

\subsubsection{Sequential Relation Identification}

\noindent Table~\ref{tab:relation-judgement} shows the experimental results for sequential relation classification, from which we find that the simple PMI baseline can achieve a good performance. Indeed, due to the imbalance of positive and negative test examples, PMI baseline chooses a threshold to classify all test examples as positive, and get a recall of 1. Four different classifiers using all the features in Table~\ref{tab:features} achieve poor results, and only the NB classifier achieves higher performance than the baseline method. We explored all combinations of four kinds of features to find the best feature set for each classifier. The NB classifier achieves the best performance with the accuracy of 77.6\% and the F1 score of 85.7\% .  

\begin{table*} [tbp]\small
	\centering
	\begin{tabular*}{0.62\textwidth}{l|c c c c c}
		\toprule
		\textbf{Features} & \textbf{Classifier} & \textbf{Accuracy} & \textbf{Precision} & \textbf{Recall} & \textbf{F1} \\
		\hline
		PMI & - & 71.9 & 71.9 & 100.0 & 83.7 \\
		\hline
		\multirow {4}{*}{All Features}& NB  & \textbf{76.3} & 78.4 & \textbf{92.4} & \textbf{84.8}  \\
		&LR  & 69.0 & 79.5 & 76.5 & 77.9  \\
		&MLP  & 68.3 & 84.1 & 69.2 & 75.6  \\
		&SVM  & 52.3 & \textbf{84.9} & 40.9 & 55.1  \\
		\hline
		Ratio+Association & NB & \textbf{77.6} &   78.9 & \textbf{93.9} &    \textbf{85.7}  \\
		Ratio &LR  & 77.0 &    80.0 & 90.7 & 85.0  \\
		Association &MLP  & 74.7 & \textbf{80.8} &    85.2 & 82.9  \\
		Ratio &SVM  & 76.5 &   78.9 & 91.9 & 84.9   \\
		\bottomrule
	\end{tabular*}
	\caption{Sequential relation classification results. Baseline result is given at the top row. Results of each classifier with all four kinds of features in Table~\ref{tab:features} are in the middle. Results of each classifier with the best feature combinations are given at the bottom.}
	\label{tab:relation-judgement}
\end{table*}



\begin{table*} [tbp]\small
	\centering
	\begin{tabular*}{0.65\textwidth}{l|c c c c c c}
		\toprule
		\textbf{Features}&\textbf{Classifier} &\textbf{Accuracy} & \textbf{Precision} & \textbf{Recall} & \textbf{F1} \\
		\hline
	    Preceding	Assumption & - & 86.1 & 86.6 & 99.3 & 92.5 \\
		\hline
		\multirow {4}{*}{All Features} & NB  & 80.3 & 89.1 & 88.0 & 88.5  \\
		& LR  & 64.2 & 89.4 & 66.3 & 76.1  \\
		& MLP  & 78.7 & \textbf{90.3} & 84.4 & 87.2 \\
		& SVM & \textbf{86.4} & 86.6 & \textbf{99.7} & \textbf{92.7}  \\
		\hline
		Association & NB  & 86.2 &  86.3 & \textbf{99.9} &    92.6  \\
		Ratio+Association & LR  & 71.3 &    86.1 & 79.6 & 82.6  \\
		All Features & MLP  & 78.7 &    \textbf{90.3} &    84.4 & 87.2  \\
		Association+Context & SVM & \textbf{87.0} &    87.7 & 98.8 & \textbf{92.9}  \\
		\bottomrule
	\end{tabular*}
\caption{Sequential direction classification results. Baseline result is given at the top row. Results of each classifier with all four kinds of features in Table~\ref{tab:features} are in the middle. Results of each classifier with the best feature combinations are given at the bottom.}
	\label{tab:direction-judgement}
	
\end{table*}


Table~\ref{tab:direction-judgement} shows the experimental results for sequential direction classification, from which we find that the \textbf{Preceding Assumption} is a very strong baseline for direction classification, and achieves an accuracy of 86.1\% and a F1 score of 92.5\%. Four classifiers with all features in Table~\ref{tab:features} achieve poor results, and only the SVM achieves higher performance than the baseline method. We explored all combinations of four kinds of features, to find the best feature set for different classifiers. Still, the SVM classifier achieves the best performance with an accuracy of 87.0\% and a F1 score of 92.9\%, using the association and context based features.




\subsubsection{Causal Relation Extraction}

\begin{table} [!tb]\small
	\centering
	\begin{tabular*}{0.49\textwidth}{lp{1cm}p{1cm}p{0.6cm}p{0.5cm}}
		\toprule
		\textbf{Methods} &\textbf{Accuracy} & \textbf{Precision} & \textbf{Recall} & \textbf{F1} \\
		\hline
		rule-based approach & 62.1 & \textbf{93.7} & 59.3 & 72.6  \\
		Bert-based approach & \textbf{85.1} & 90.6 & \textbf{77.6} & \textbf{83.6}  \\
		\bottomrule
	\end{tabular*}
\caption{Causality extraction performance.}
	\label{tab:causality_extraction}
	
\end{table}

\noindent Table~\ref{tab:causality_extraction} shows the experimental results for causal relation extraction, from which we find Bert-based approach dramatically outperforms rule-based approach. This is mainly because the BERT model can obtain general language knowledge from pre-training, and then our annotated data can be used to fine-tune the model to extract the causal relation. Rule-based approach can achieve better precision score but worse recall score, because manually constructed rules can hardly cover the whole linguistic phenomenons. 

\subsubsection{Script Event Prediction}
\begin{table}[!tb] \small
	\centering
	\begin{tabular} {l c} 
		\toprule
		\textbf{Methods}& \textbf{Accuracy}  \\
		\midrule
		Random & 20.00 \\
		PMI \cite{chambers2008} & 30.52 \\
		Bigram \cite{jans2012skip}& 29.67 \\
		Word2vec \cite{mikolov2013distributed}& 42.23 \\
		DeepWalk  \cite{perozzi2014deepwalk} & 43.01 \\
		EventComp \cite{MarkGW-AAAI16}& 49.57  \\
		PairLSTM \cite{wang2017integrating}& 50.83 \\
		\midrule
		SGNN-attention (without attention) & 51.56  \\
		SGNN (ours) & \textbf{52.45}  \\
		SGNN+PairLSTM & 52.71  \\
		SGNN+EventComp & 54.15  \\
		SGNN+EventComp+PairLSTM & \textbf{54.93} \\
		\bottomrule
	\end{tabular}
	\caption{Results of script event prediction accuracy (\%) on the test set. SGNN-attention is the SGNN model without attention mechanism. Differences between SGNN and all baseline methods are significant ($p < 0.01$) using t-test, except SGNN and PairLSTM ($p = 0.246$).}
	\label{tab:final}
\end{table}

\noindent Experimental results are shown in Table \ref{tab:final}, from which we can make the following observations.

(1) Word2vec, DeepWalk and other neural network-based models (EventComp, PairLSTM, SGNN) achieve significantly better results than the counting-based PMI and Bigram models. The main reason is that learning low dimensional dense embeddings for events is more effective than sparse feature representations for script event prediction.

(2) Comparison between ``Word2vec'' and ``DeepWalk'', and between ``EventComp, PairLSTM'' and ``SGNN'' show that graph-based models achieve better performance than pair-based and chain-based models. This confirms our assumption that the event graph structure is more effective than event pairs and chains, and can provide more abundant event interactions information for script event prediction.

(3) Comparison between ``SGNN-attention'' and ``SGNN'' shows the attention mechanism can effectively improve the performance of SGNN. This indicates that different context events have different significance for choosing the correct subsequent event.

(4) SGNN achieves the best script event prediction performance of 52.45\%, which is 3.2\% improvement over the best baseline model (PairLSTM).

We also experimented with combinations of different models, to observe whether different models have complementary effects to each other. We find that SGNN+EventComp boosts the SGNN performance from 52.45\% to 54.15\%. This shows that they can benefit from each other. Nevertheless, SGNN+PairLSTM can only boost the SGNN performance from 52.45\% to 52.71\%. This is because the difference between SGNN and PairLSTM is not significant, which shows that they may learn similar event representations but SGNN learns in a better way. The combination of SGNN, EventComp and PairLSTM achieves the best performance of 54.93\%. This is mainly because pair structure, chain structure and graph structure each has its own advantages and they can complement each other.

\begin{figure} [t] 
	\centering
	\includegraphics[width=1\columnwidth]{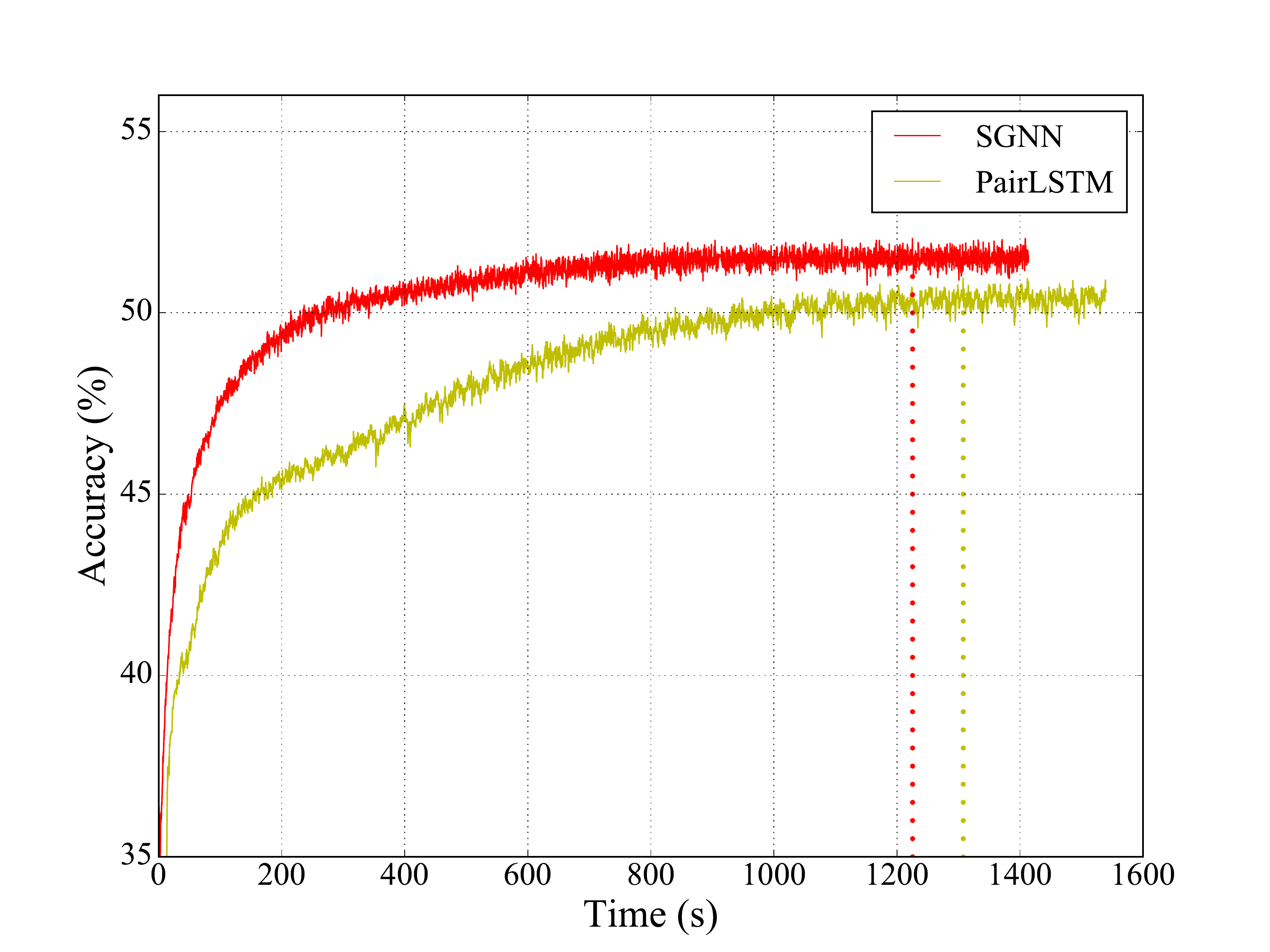}
	\caption{Learning curves on development set of SGNN and PairLSTM models, using the same learning rate and batch size.}
	\label{fig:time}
\end{figure}

The learning curve (accuracy with time) of SGNN and PairLSTM is shown in Figure \ref{fig:time}. We find that SGNN quickly reaches a stable high accuracy, and outperforms PairLSTM from start to the end. This demonstrates the advantages of SGNN over PairLSTM model.

\section{Related Work}
\noindent The most relevant research area with ELG is script learning. The use of scripts in AI dates back to the 1970s~\cite{schank1977scripts,mooney1985learning}. In this study, \emph{scripts} are an influential early encoding of situation-specific world event. In recent years, a growing body of research has investigated statistical script learning. Chambers et al.,~\shortcite{chambers2008} proposed unsupervised induction of \textit{narrative event chains} from raw newswire text, with \textit{narrative cloze} as the evaluation metric. Jans et al.,~\shortcite{jans2012skip} used bigram model to explicitly model the temporal order of event pairs. However, they all utilized a very simple representation of event as the form of (\textit{verb, dependency}). To overcome the drawback of this event representation, Pichotta and Mooney~\cite{pichotta2014statistical} presented an approach that employed events with multiple arguments. 

There have been a number of recent neural models for script learning. Pichotta and Mooney~\shortcite{pichotta2016learning} showed that LSTM-based event sequence model outperformed previous co-occurrence-based methods for event prediction. 
Mark and Clark~\shortcite{MarkGW-AAAI16} described a feedforward neural network which composed verbs and arguments into low-dimensional vectors, evaluating on a multiple-choice version of the Narrative Cloze task. 
Wang et al.,\shortcite{wang2017integrating} integrated event order information and pairwise event relations together by calculating pairwise event relatedness scores using the LSTM hidden states as event representations.

Script learning is similar to ELG in concepts. However, script learning usually extracts event chains without considering their temporal orders and causal relations.  ELG aims to organize event evolutionary patterns into a commonsense knowledge base, which is the biggest difference with script learning.

\section{Conclusion}

\noindent In this paper, we present an Event Logic Graph (ELG), which can reveal the evolutionary patterns and development logics of real world events. We also propose a framework to construct ELG from large-scale unstructured texts, and use the ELG to improve the performance of script event prediction. All techniques used in the ELG are language-independent. Hence, we can easily construct other language versions of ELG.

\section*{Acknowledgments}

\noindent This work is supported by the National Key Basic Research Program of China via grant 2014CB340503 and the National Natural Science Foundation of China (NSFC) via grants 61472107 and 61702137. The authors would like to thank the anonymous reviewers.

\bibliography{acl2019}
\bibliographystyle{acl_natbib}

\end{document}